\documentclass[10pt,twocolumn,letterpaper]{article}

\usepackage{cvpr}              

\usepackage{graphicx}
\usepackage{amsmath}
\usepackage{amssymb}
\usepackage{pifont}
\usepackage{booktabs}
\usepackage{stfloats}
\usepackage{xcolor}

\usepackage[pagebackref,breaklinks,colorlinks]{hyperref}

\usepackage[capitalize]{cleveref}
\crefname{section}{Sec.}{Secs.}
\Crefname{section}{Section}{Sections}
\Crefname{table}{Table}{Tables}
\crefname{table}{Tab.}{Tabs.}

\def\cvprPaperID{31} 
\def\confName{CVPR}
\def\confYear{2024}

\begin{document}

\title{Sparse multi-view hand-object reconstruction for unseen environments}

\author{Yik Lung Pang\textsuperscript{1}, Changjae Oh\textsuperscript{1}, Andrea Cavallaro\textsuperscript{1,2,3} \\
\textsuperscript{1}Centre for Intelligent Sensing, Queen Mary University of London \\ \textsuperscript{2}Idiap Research Institute \textsuperscript{3}École Polytechnique Fédérale de Lausanne\\
{\tt\small \{y.l.pang, c.oh\}@qmul.ac.uk, a.cavallaro@idiap.ch}
}

\maketitle

\begin{abstract}

     Recent works in hand-object reconstruction mainly focus on the single-view and dense multi-view settings. On the one hand, single-view methods can leverage learned shape priors to generalise to unseen objects but are prone to inaccuracies due to occlusions. On the other hand, dense multi-view methods are very accurate but cannot easily adapt to unseen objects without further data collection. In contrast, sparse multi-view methods can take advantage of the additional views to tackle occlusion, while keeping the computational cost low compared to dense multi-view methods. In this paper, we consider the problem of hand-object reconstruction with unseen objects in the sparse multi-view setting. Given multiple RGB images of the hand and object captured at the same time, our model SVHO combines the predictions from each view into a unified reconstruction without optimisation across views. We train our model on a synthetic hand-object dataset and evaluate directly on a real world recorded hand-object dataset with unseen objects. We show that while reconstruction of unseen hands and objects from RGB is challenging, additional views can help improve the reconstruction quality.
     
\end{abstract}
\begin{figure*}[ht!]
    \centering

    \includegraphics[width=0.9\textwidth,trim={0cm 1cm 2cm 1cm},clip]{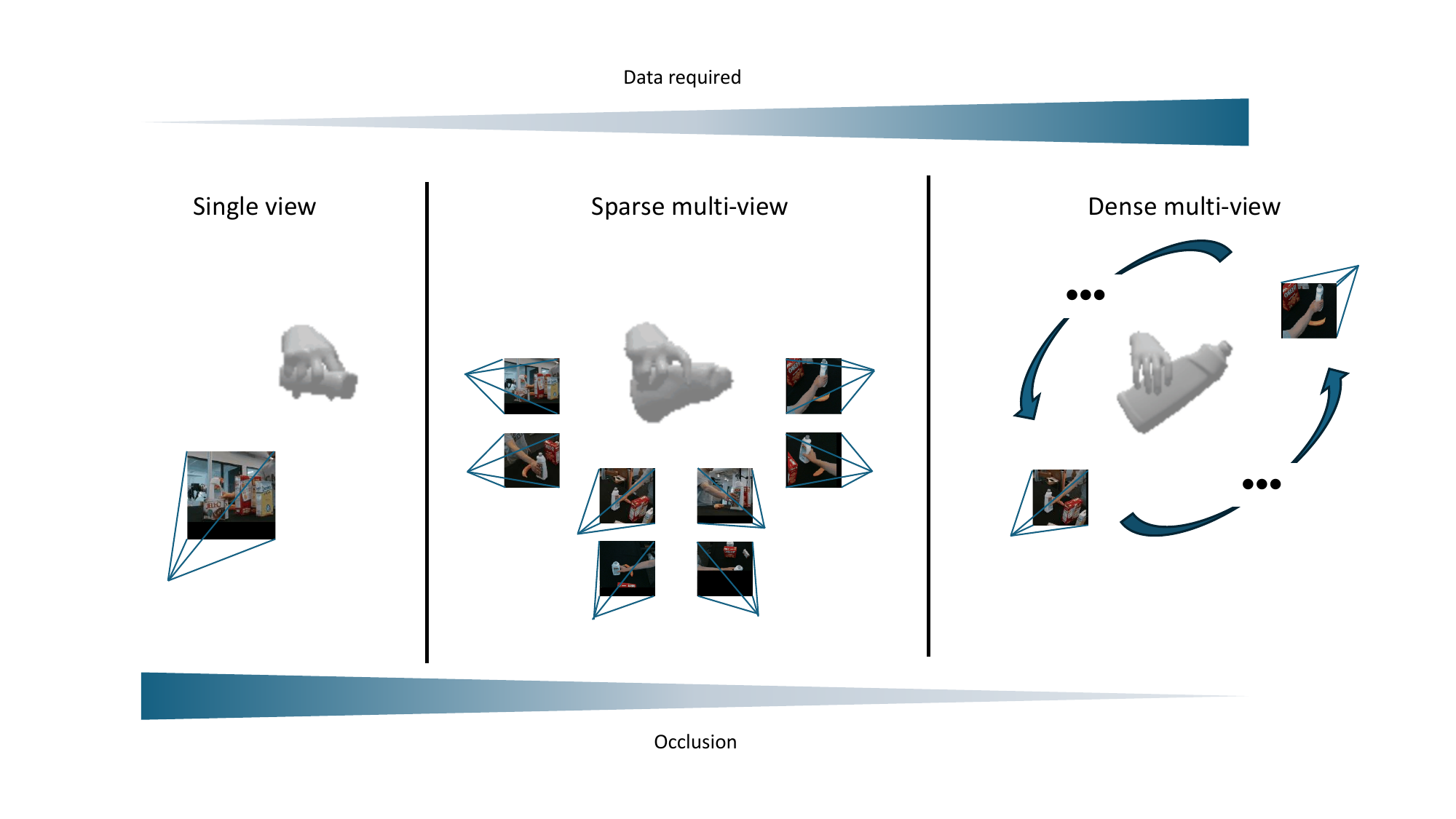}
    
    \caption{Single-view methods suffer from occlusion, while dense multi-view methods require a large amount of collected images. We propose to use sparse multi-view input to improve the reconstruction quality while keeping the data requirements low.
    }
    \label{fig:overview}
    \vspace{-5pt}
\end{figure*}
\section{Introduction}
\label{sec:intro}
Reconstructing hand and object shapes from visual input has a lot of applications ranging from AR/VR to robotics~\cite{yang2021reactive,sanchez2020benchmark,rosenberger2020object,pang2021towards}. For example, in a human-to-robot handover scenario, the retrieved shapes help guide the robot to estimate possible grasps on the object while avoiding contact with the human hand. In this case, the accuracy of the reconstruction method is important, as a poorly estimated object shape will cause the grasp to fail, while a poorly estimated hand shape can cause injury to the human if the robot is unsure where the human hand is. Moreover, if the human attempts to handover an unseen object to the robot, the robot should be able to adapt to this new object on the fly without additional data collection to ensure the naturalness of the interaction. 

Joint hand-object shape reconstruction from RGB has seen impressive progress in recent years~\cite{hasson19_obman,tse2022collaborative,chen2022alignsdf,chen2021joint,ye2022s,karunratanakul2020grasping,chen2023gsdf,choi2024handnerf} due to the availability of high-quality parametric hand models~\cite{romero2022embodied} and neural implicit representations~\cite{park2019deepsdf} for modelling complex 3D shapes. However, these methods are still rarely applied in the human-to-robot handover scenario either due to poor reconstruction quality or large data collection requirements.

Recent hand-object shape reconstruction methods can be separated into two categories, namely single-view or dense multi-view methods. Single-view methods can reconstruct hand-object shapes in a single forward pass by leveraging learned 3D shape priors from large hand-object interaction datasets~\cite{hasson19_obman,tse2022collaborative,chen2022alignsdf,chen2021joint,ye2022s,karunratanakul2020grasping,chen2023gsdf,choi2024handnerf}. However, predicting the shape of unknown hands and objects from a single-view is challenging as the diversity of the hand poses and object shapes is vast. Moreover, when the object or hand is occluded or under motion blur, the reconstruction quality suffers greatly~\cite{ye2022s,chen2022alignsdf,chen2023gsdf}.

Using multiple views can alleviate the problem of occlusion by providing redundancy in the input. Neural implicit shape representation enables detailed reconstruction of unknown shapes from densely captured images of the object from multiple viewpoints~\cite{hampali2022hand,swamy2023showme,qu2023novel,ye2023diffusion}. However, the reconstruction suffers where the object is occluded, which is common in the hand-object interaction scenario. Moreover, obtaining and optimising on dense multi-view images is time consuming and is not suitable in scenarios where the model has to adapt quickly to an unseen object.

Sparse multi-view methods provide a balanced approach between single-view and dense multi-view methods but has not been investigated in the hand-object reconstruction task. Compared to single-view methods, the additional views can help alleviate issues caused by occlusion or challenging poses. On the other hand, the data requirement is lower compared to dense multi-view methods, which enables applications where the model has to reconstruct unseen objects without further data collection or training.

In this paper, we propose SVHO, a sparse multi-view method for hand-object reconstruction. Our proposed method takes as input the RGB images and the corresponding global hand poses from each view. The model predicts the hand and object shapes independently from each view and combines them to form a final reconstruction. We train our model entirely on the synthetic dataset ObMan~\cite{hasson19_obman} and evaluate on the real-world recorded dataset DexYCB~\cite{chao2021dexycb} with unseen objects.

In summary, our contributions are:
\begin{itemize}
    \item We propose a sparse multi-view setting for hand-object reconstruction
    \item We analyse our proposed sparse multi-view reconstruction method on up to 8 input views of unseen objects
\end{itemize}

\section{Related works}
\label{sec:related_works}

\subsection{3D shape representations}
Early works in hand-object shape estimation focused on using triangular meshes as a representation for the hand and object shape~\cite{hasson19_obman,tse2022collaborative}. The MANO model~\cite{romero2022embodied} provided a low dimensional representation that can be decoded back to a complete hand mesh efficiently. On the other hand, predicting meshes for unknown objects is challenging, especially for non genus 0 shapes~\cite{groueix2018papier}. Neural implicit shape representations~\cite{park2019deepsdf,chen2019learning} have recently emerged as a popular choice for representing shapes in 3D due to their versatile capabilities of representing complex shapes, and have been adopted in hand object reconstruction works~\cite{chen2022alignsdf,chen2021joint,ye2022s,karunratanakul2020grasping,chen2023gsdf}.

\subsection{Single-view hand-object reconstruction}

Works on hand-object shape reconstruction mainly follow the encoder-decoder architecture. An image encoder encodes the input image to the latent space, and then a 3D shape decoder maps the latent vectors to triangular meshes or signed distance fields (SDFs). A dual branch network was proposed in~\cite{hasson19_obman}, each with a ResNet-18~\cite{he2016deep} encoder pre-trained on ImageNet~\cite{russakovsky2015imagenet} to estimate features for the hand and object respectively. The hand features are passed through a fully connected layer to estimate the shape and pose parameters for the MANO~\cite{romero2022embodied} hand model and scale and translation parameters to help determine the size of the object. The object features are passed to AtlasNet~\cite{groueix2018papier} to estimate the object shape as an atlas of mesh surfaces in normalised vertices. A contact loss was also proposed to encourage contact between the hand and the object. However, there is no exchange of information between the hand and object shape estimation branch, which both contains important information for each other's task (e.g. estimating contact surfaces).

Later works explored the idea of sharing features between the hand and the object to improve the reconstruction quality. Relevant information between the hand and object can be shared via attention-guided graph convolution to improve both the estimation of the hand and object shape in an iterative manner~\cite{tse2022collaborative}. However, the approach requires calculating the attention value of a fully connected graph formed from the hand and object mesh in multiple iterations, which can be time consuming. The speed and accuracy of the graph convolution architecture can be improved by introducing a dense mutual attention mechanism~\cite{wang2023interacting}. This removes the need to optimise the shapes in iterative steps. However, the features are shared in the global latent space, which can be difficult to optimise.

Several works have improved reconstruction results by estimating local shape features instead of a single global latent feature~\cite{ye2022s,chen2022alignsdf,chen2023gsdf}. The estimation of the object shape can be conditioned on the estimated hand shape~\cite{ye2022s}. After estimating the hand shape using an off-the-shelf estimator, the object is reconstructed in the hand-wrist frame. Each query 3D position is projected back to the 2D image to retrieve the corresponding image features and the distance to each hand joint is encoded as additional features for the reconstruction of the object. Similar projection based feature retrieval was also proposed in~\cite{chen2022alignsdf,chen2023gsdf}, while also estimating object poses in addition to hand poses as condition to the shape reconstruction process. Most recently, a NeRF-based hand-object reconstruction model was proposed for single-view reconstruction which can be trained directly from multi-view images using photometric loss and does not require 3D ground-truth shape as supervision~\cite{choi2024handnerf}.

\subsection{Multi-view hand-object reconstruction}
Multi-view reconstruction of hand and object relies on the optimisation of the predicted shape and appearance based on densely collected images with predicted hand global pose~\cite{hampali2022hand,swamy2023showme,qu2023novel,ye2023diffusion}. NeRF~\cite{mildenhall2021nerf} based methods were proposed to represent the hand and object shape~\cite{hampali2022hand,swamy2023showme}. The pose and shape of the hand and object are optimised by minimising the photometric loss between the rendered colour and the observed colour. These methods require a prerecorded video of the interaction and are not able to infer the shape in a new view. Moreover, the occluded parts of the object cannot be reconstructed as it is never observed by the camera. A diffusion-based prior was proposed to estimate the geometry of the parts occluded in the video~\cite{ye2023diffusion}. A model was proposed to generalise to sparse novel views at test time~\cite{qu2023novel}. However, the model has to be trained offline first with sparse view images of the same hand and object separately.

\section{Proposed method}
We aim to reconstruct the shape of the hand and the shape of the hand-held object from sparse multi-view images. We first predict the hand and object shape from each camera view independently in the canonical coordinate space. We then combine them to obtain a single final output.

We train autoencoders to encode hand and object shapes independently in the canonical coordinate space. We train a Patchwise VQ-VAE (P-VQ-VAE)~\cite{autosdf2022} to autoencode the hand and object into 3D discrete latent cubes with corresponding shape codebooks. This allows us to treat the shape estimation problem as a classification problem with the classes being the codebook indices.

During test time, we obtain 2D features from the input image. We then form a 3D feature grid by projecting the 3D points in the canonical coordinate space to the image space using the global hand pose. We reconstruct the hand and object shapes in the canonical coordinate space and combine the classification probabilities of the different views. The final mesh is obtained using the trained decoder and the marching cubes algorithm~\cite{lorensen1998marching}.

\subsection{Autoencoding hands and objects}

\begin{figure}[t!]
    \centering
    \begin{tabular}{c}
        \includegraphics[width=0.9\columnwidth]{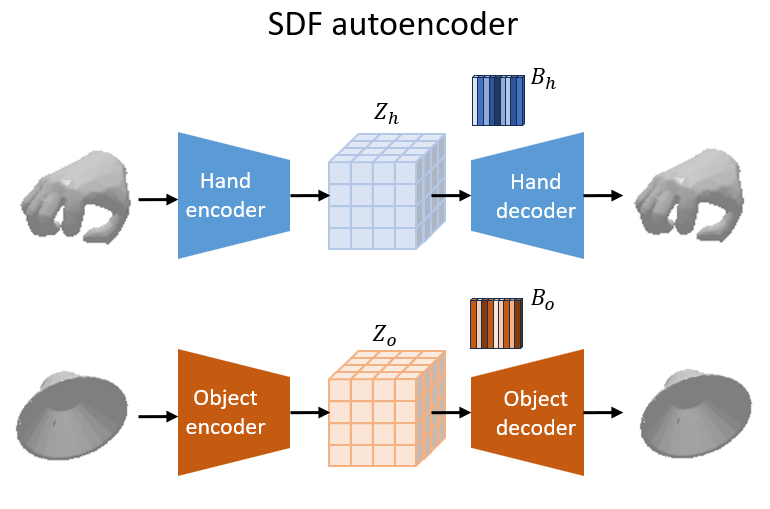}\\
    \end{tabular}
    \caption{We first encode hand and object shape independently using Patchwise VQ-VAE (P-VQ-VAE). This provides a compact representation to train our hand object shape prior.
    }
    \label{fig:autoencode}
    \vspace{-5pt}
\end{figure}

We train autoencoders to map the 3D shape of the hand and the object into 3D discrete latent cubes ${Z}_h \in \mathbb{N}^{8 \times 8 \times 8}$ and ${Z}_o \in \mathbb{N}^{8 \times 8 \times 8}$, respectively. We use P-VQ-VAE~\cite{autosdf2022} as the encoder architecture. The input SDF $\mathbb{R}^{128 \times 128 \times 128}$ is first divided into $8 \times 8 \times 8 = 512$ patches of $\mathbb{R}^{16 \times 16 \times 16}$. Each patch is encoded individually with 3D convolutional layers into a latent vector $z_e \in \mathbb{R}^{V}$ which combine to give the continuous latent cube of $Z' \in \mathbb{R}^{8 \times 8 \times 8 \times V}$. We then learn a codebook $B \in \mathbb{R}^{K \times V}$ with $K$ elements of $e_k \in \mathbb{R}^{V}$, and discretize the continuous latent cube by mapping each patch's latent vector $z_e \in \mathbb{R}^{V}$ to its nearest element $e_i$ in the codebook~\cite{van2017neural}.
\begin{equation}
    z = e_i, \textrm{ where } i = \textrm{argmin}_k{||z_e - e_k||}_2,
\end{equation}
where $i$ is the index of the element $e_i$ in the codebook nearest to the encoder output $z_e$, and $z\in \mathbb{R}^{V}$ is the mapped latent code. We save the extracted index $i$ for each patch and reshape them back into the 3D cube $Z \in \mathbb{N}^{8 \times 8 \times 8}$. Note that we omit the subscript $h$ and $o$ as the same process is repeated for the hand and object.

By learning the codebook $B$ we obtain a compact discrete latent representation of the hand and object shape. This representation preserves the local shape information of the hand and object as each latent code corresponds to the local shape input only. Together, the latent cube is decoded jointly to ensure the consistency of the global shape. The continuous codes $Z' \in \mathbb{R}^{8 \times 8 \times 8 \times V}$ are first retrieved using the codebook $B$ and discrete latent cube $Z$ and enhanced with 3D convolutional layers across neighbouring patches to ensure spatial consistency.

We learn a decoder $D$ to map latent codes $e$ and 3D position $p = (x,y,z)$ to predicted SDF value $\hat{s}$. We use a 5-layer MLP decoder to be consistent with previous works~\cite{chen2023gsdf,chen2022alignsdf}. To retrieve the latent code $e$, the mixed codes are upsampled using trilinear interpolation to $\mathbb{R}^{128 \times 128 \times 128 \times V}$, and the corresponding code is retrieved using the query 3D position $p$. We first obtain the corresponding latent cube index $(i,j,k) = M(p)$ with the indexing function $M$. We then retrieve $e = Z'_{ijk}$ as the latent code for position $p$. We optimise the loss $L_{AE}$ to train the autoencoders:
\begin{equation}
    L_{AE} = ||s-\hat{s}|| + ||sg[z_e]-e||^2_2 + \beta||z_e-sg[e]||^2_2,
\end{equation}
where $s$ and $\hat{s}$ are the ground-truth and predicted signed distance field (SDF) values respectively, $sg$ is the stop gradient operator, and $e$ is the nearest element in codebook $B$ to encoder output $z_e$. The first term minimises the difference between the reconstructed and input shape, while the second and third terms are the vector quantisation loss and commitment loss~\cite{van2017neural}.

\subsection{Reconstruction from multi-view images }
\begin{figure*}[t!]
    \centering
    \begin{tabular}{c}
    \includegraphics[width=1.8\columnwidth]{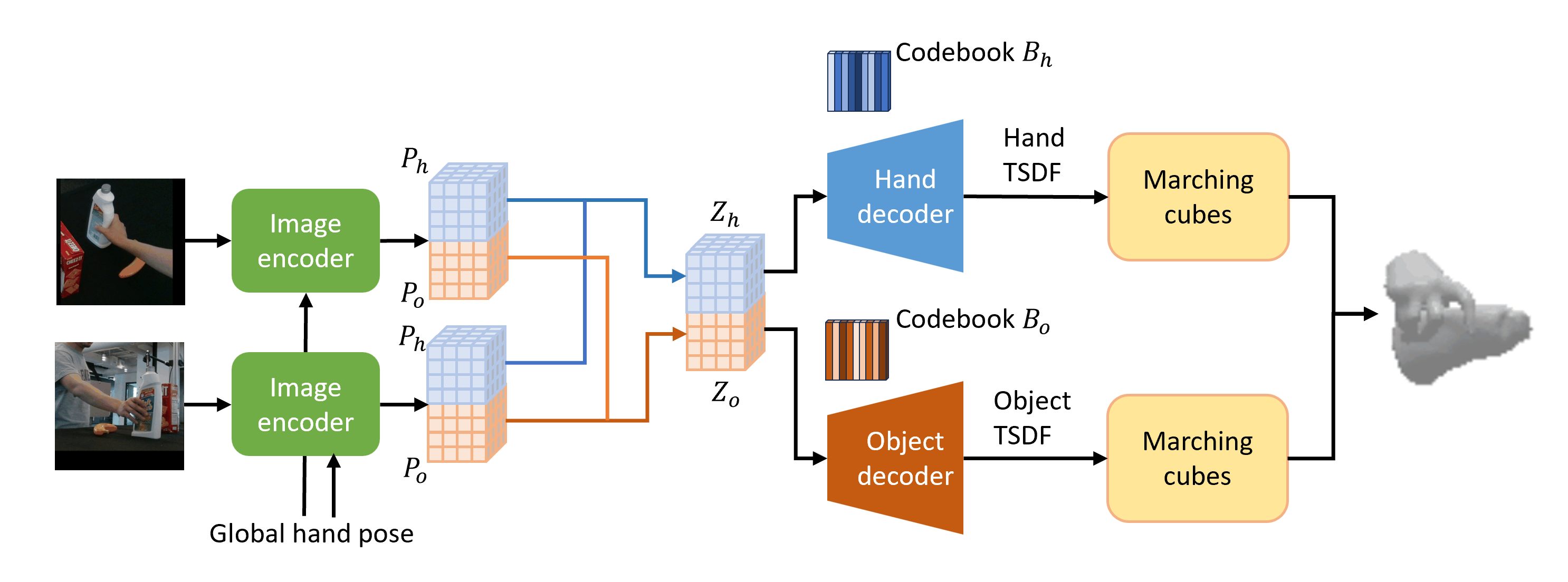}\\
    \end{tabular}
    \caption{Our pipeline for hand object shape reconstruction from multi-view images. Predicted probabilities from individual views are averaged to get the final prediction.
    }
    \label{fig:image_recon}
    \vspace{-5pt}
\end{figure*}

After the autoencoders are trained, we run the encoder on the training set to extract the training pairs $\{I, Z_h, Z_o, \pi\}$, where $I$ is the RGB image and $\pi$ is the ground-truth transformation from the camera coordinate frame to the hand-wrist coordinate frame.

We first train our model on the single-view reconstruction task. We use ResNet-18~\cite{he2016deep} to encode the input image into 2D features. Spatially aligned 3D features are retrieved by projecting the query 3D point back to the 2D image space using the hand pose. We treat the reconstruction task as a classification task with the codebook indices as the classes~\cite{autosdf2022}. We use 3D convolutional layers to process the spatially aligned 3D features and output $\hat{P} \in \mathbb{R}^{8 \times 8 \times 8 \times K}$ which corresponds to the predicted probabilities of each codebook index at each location in 3D space. We train the model using a weighted cross-entropy where a weight of $0.25$ is given to the index corresponding to an empty space and a weight of $0.75$ is given to all other indices equally as the dataset is biased towards empty space.

To reconstruct the hand and object shapes from multiple views $C$ with input images $I = \{I_1, ..., I_{|C|}\}$, we first obtain independent reconstructions from each camera view $c$. We aggregate the results across different views as $P$ by averaging the predicted probabilities of the selected views. 
\begin{equation}
    P = \frac{1}{|C'|}{\sum_{c \in C'}}{\hat{P_c}},
\end{equation}
where $C' \subseteq C$ is the subset of views selected for reconstruction. After $P$ is obtained, we select the codebook index $i$ at each location with the highest predicted probability to form the predicted latent cube $\hat{Z}$. We retrieve the continuous latent cube $\hat{Z'}$ by indexing the codebook $B$ and we use the trained decoder $D$ to obtain the predicted SDF values $\hat{s}$. Finally, we obtain the reconstructed mesh $R_{C'}$ using marching cubes~\cite{lorensen1998marching} from the predicted SDF values.

\section{Experiments}
\subsection{Datasets}
\paragraph{ObMan~\cite{hasson19_obman}} We train our model on the synthetic only dataset ObMan containing 8 object categories (bottles, bowls, cans, jars,
knifes, cellphones, cameras and remote controls) from ShapeNet~\cite{chang2015shapenet}. Each sample contains an annotated hand grasp with the corresponding ground-truth hand and object mesh. We follow AlignSDF~\cite{chen2022alignsdf} and discard the samples with too many double-sided triangles, resulting in a training set of 87190 samples and a testing set of 6285 samples.
\paragraph{DexYCB~\cite{chao2021dexycb}} We use 
DexYCB as a testing dataset that includes multi-view video recordings of humans performing grasping of objects. The dataset contains 20 unique objects from the YCB-Video dataset~\cite{xiang2018posecnn}. Following AlignSDF~\cite{chen2022alignsdf}, we only consider the samples in the S0 split where the minimum distance between the hand and the object mesh is less than 5mm. For each object, we randomly sample 100 frames, where each frame contains 8 RGB images from the 8 camera views, the ground-truth hand and object mesh, and the annotated hand pose. This results in a total of 2000 sampled frames, with 16000 images.

\begin{figure*}[t!]
    \centering
    \begin{tabular}{c}
        \includegraphics[width=2.0\columnwidth,trim={0 5cm 0 1cm},clip]{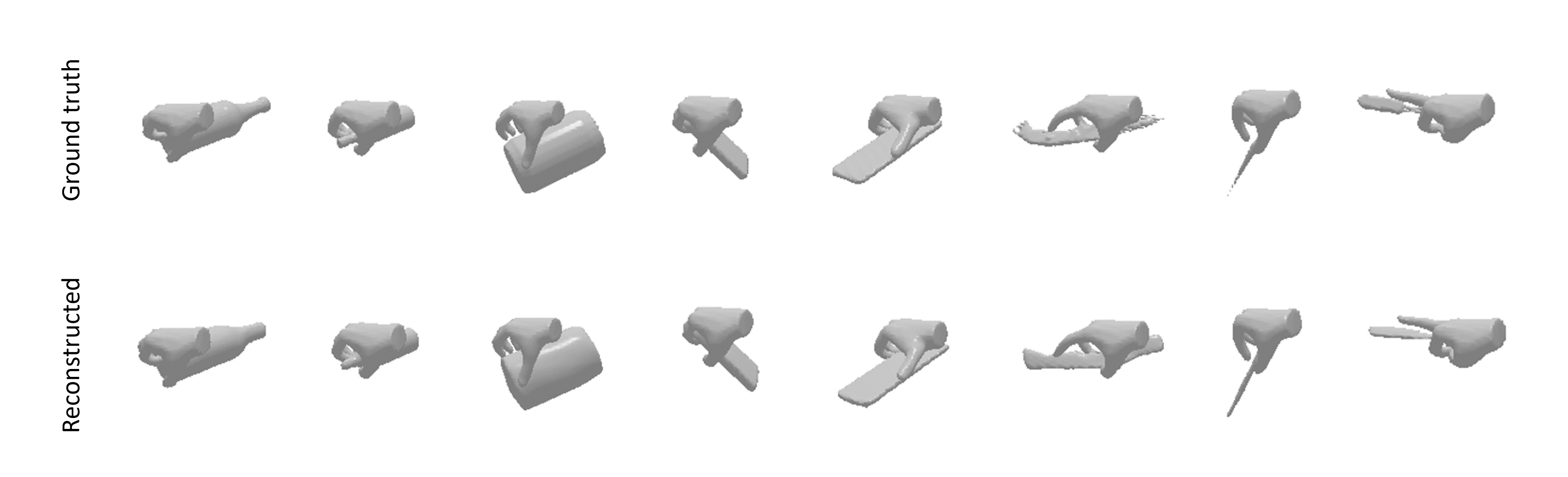}\\
        \includegraphics[width=2.0\columnwidth,trim={0 0.5cm 0 4.5cm},clip]{results_autoencoder2.png}\\
    \end{tabular}
    \caption{Autoencoder reconstruction of 3D hand and objects
    }
    \label{fig:vis_autoencoder}
    \vspace{-5pt}
\end{figure*}
\subsection{Implementation details}
Our goal is to mimic the scenarios in human-robot interactions where the robot has to adapt quickly to unseen objects grasped by humans. Therefore, we train our model entirely on the synthetic ObMan dataset, and we evaluate the performance on the DexYCB dataset where the objects used are unseen.

We first train our P-VQ-VAE to autoencode hand and objects independently on the training set of ObMan. We use a codebook size of 512 with a latent code of length 128, and we sample the hand and object SDF as a truncated SDF (TSDF) of size $128\times128\times128$. Due to instability in the codebook training, we perform codebook restart~\cite{dhariwal2020jukebox} on codes that are not used every 25 batches. We double this duration every time the coodebook is restarted. Once the model is trained, we run the encoders on the training set to obtain the latent cube $Z$ for each sample.

We then pair these sampled latent cubes $Z$ with their corresponding RGB image and global hand pose to form the training set for the image reconstruction model. Our image reconstruction model outputs the probabilities for each entry in the codebook at each location in the latent cube. We choose the predicted codebook index with the highest probability to perform the reconstruction. We retrieve the continuous latent cube $Z'$ from the codebook and use marching cubes to obtain the final mesh at the $0$ level set.

\subsection{Metrics}

Following gSDF~\cite{chen2023gsdf}, we first align the predicted and ground-truth hand mesh with the global hand pose. The object meshes are transformed also using the global hand pose. We use Chamfer distance ($cm^2$) and F-score to evaluate the reconstruction quality of the predicted hand and object meshes.

\begin{table}[t]
\centering
\small
\renewcommand{\arraystretch}{1.2}
\setlength\tabcolsep{2pt}
\caption{Results for autoencoding hand and object shapes ($CD$ - Chamfer distance; $FS$ - F-score)}

\begin{tabular}{cccccc}

\specialrule{1.2pt}{3pt}{0.6pt}

$\textit{CD}_h \downarrow$ & 
 $\textit{FS}_h@1 \uparrow$ &
 $\textit{FS}_h@5 \uparrow$ &
 $\textit{CD}_o \downarrow$ &
 $\textit{FS}_o@5 \uparrow$ &
 $\textit{FS}_o@10 \uparrow$ \\
\midrule
 0.015 & 0.798 & 0.999 & 0.437 & 0.822 & 0.953 \\



\specialrule{1.2pt}{3pt}{1pt}


\end{tabular}
\label{tab:scores}
\vspace{-10pt}
\end{table}

\subsection{Results}

\subsubsection{3D shape autoencoding}
We evaluate our P-VQ-VAE autoencoder on the testing set of ObMan. Our model takes as input the ground-truth TSDF and outputs the reconstructed TSDF of the same size. We found that our autoencoder can accurately reconstruct hands and objects of various poses and shapes (Table~\ref{tab:scores}).

Figure~\ref{fig:vis_autoencoder} shows example visualisations of hand and object reconstructions from the object classes bottles, can, remote and knife. Since the latent codes extracted by the shape encoder will act as a ground-truth for the image reconstruction method, it is important to ensure the quality of the autoencoder's reconstruction as it acts as the upperbound for our image reconstruction model.

\begin{figure}[h!]
    \centering
    \begin{subfigure}{\columnwidth}
    \includegraphics[width=.9\columnwidth,trim={-1.5cm 0 0.5cm 0},clip]{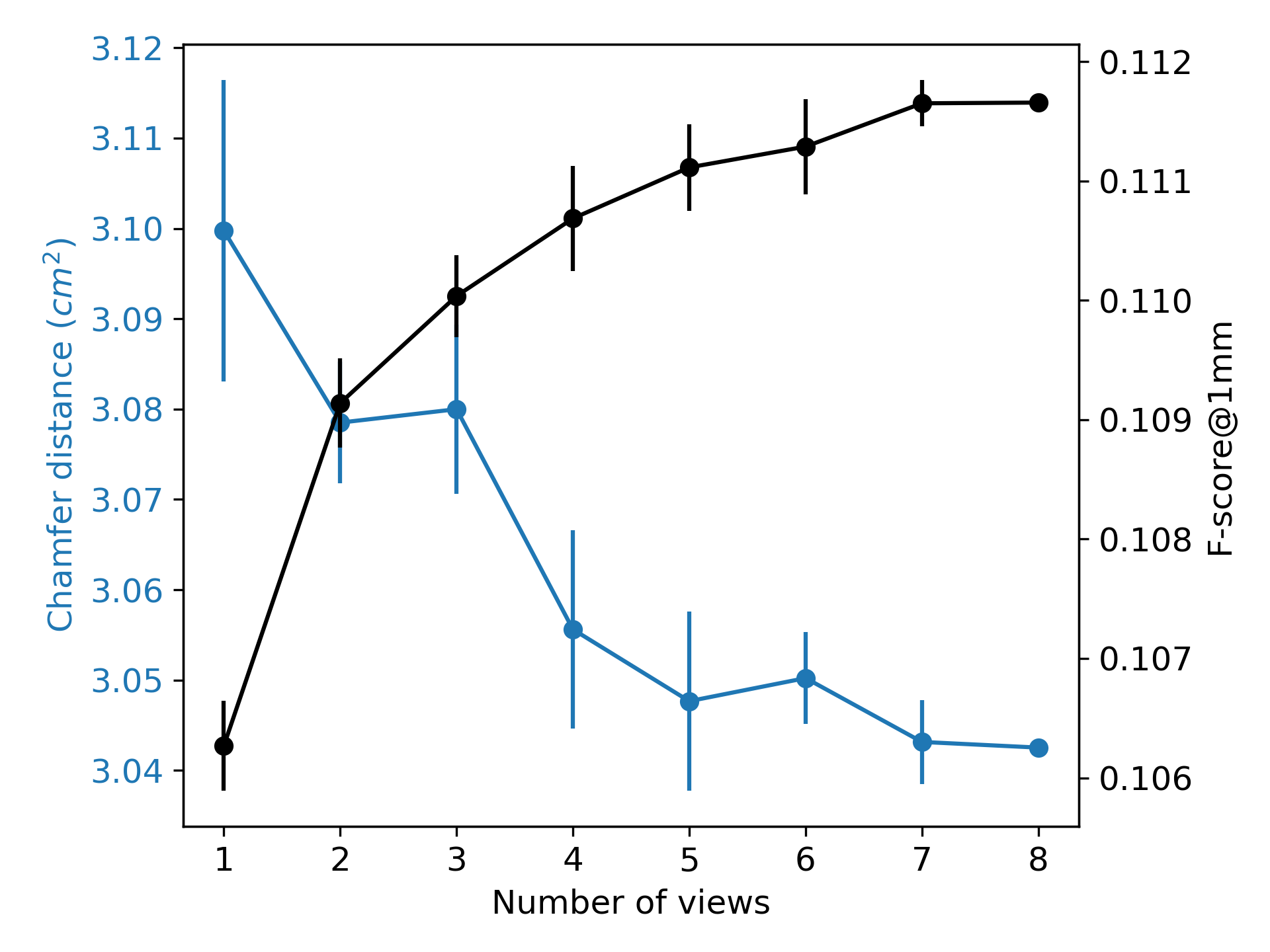}
    \caption{Hand reconstruction}
    \end{subfigure}
    \begin{subfigure}{\columnwidth}
    \includegraphics[width=.9\columnwidth,trim={-1.5cm 0 0.5cm 0},clip]{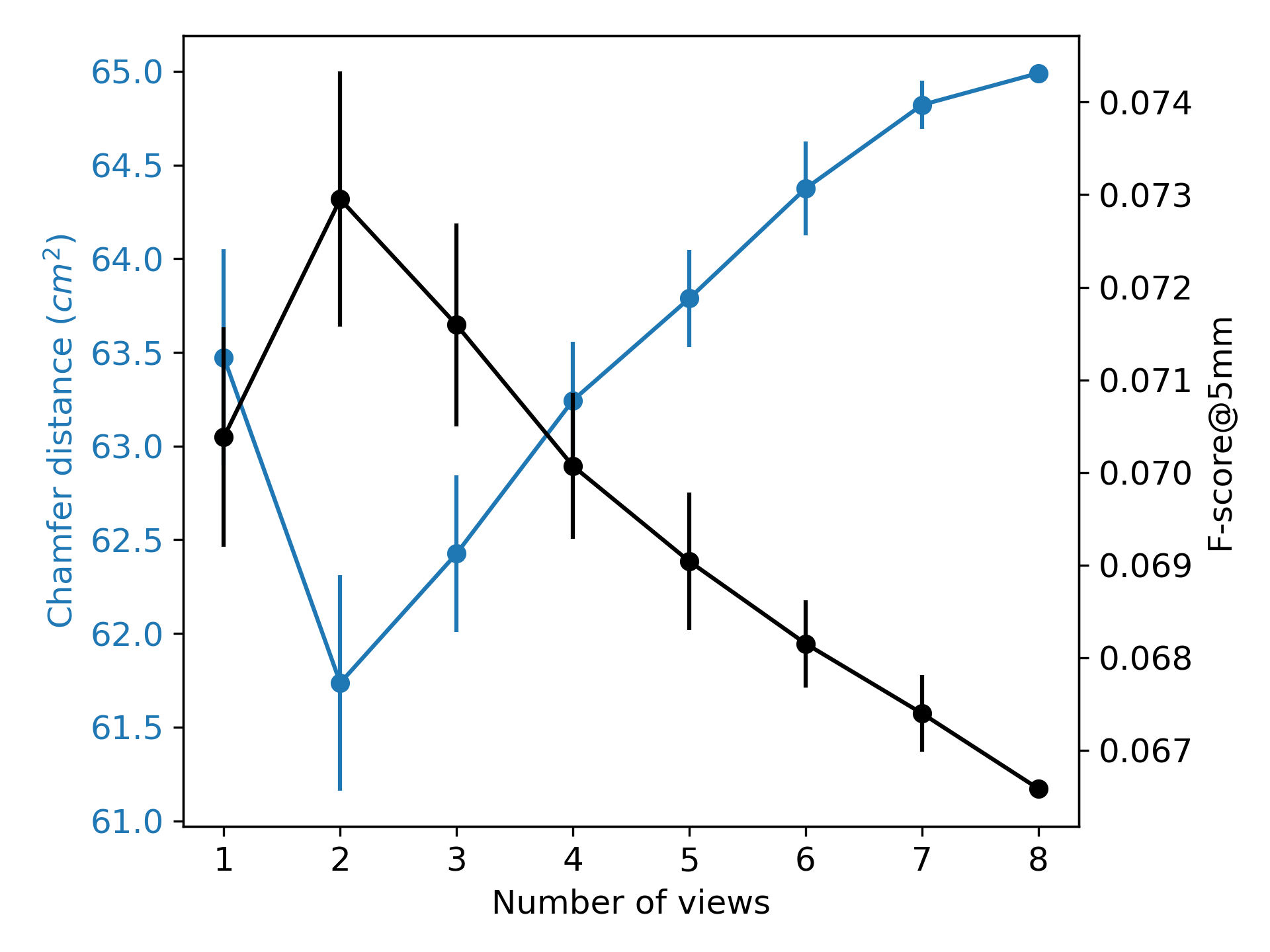}
    \caption{Object reconstruction}
    \end{subfigure}
    \caption{Average F-score and standard deviation across 6 runs for (a) hand and (b) object reconstruction when varying the number of input views
    }
    \label{fig:multi_avg}
    \vspace{-5pt}
\end{figure}
\subsubsection{Multi-view RGB reconstruction}

We evaluate our multi-view hand-object shape reconstruction model on the DexYCB dataset. To analyse the impact of the number of views on the reconstruction quality, we vary the number of input views to the model from 1 to 8. For each sample, we randomly select the corresponding number of views from the 8 available views. We repeat the experiment for each number of views 6 times and we report the mean and standard deviation across the runs.

Figure~\ref{fig:multi_avg} shows the average Chamfer distance and F-score of hand and object with varying numbers of views. For the hand, the F-score increases consistently as the number of views increases. For the object, the F-score peaks at 2 views and decreases as the number of views increases further. The Chamfer distance follows the same trend but in reverse as lower Chamfer distance is better. Since the images are centered around the wrist, the model can effectively take advantage of the multiple views to improve the features for hand reconstruction since the position of the hand in the image is known. However, as the number of views increases, other objects in the scene can distract the model from recognising the target object being grasped. Thus, more views can actually negatively impact the reconstruction quality of the object. This issue can be mitigated by introducing a hand-object segmentation model, to allow our reconstruction model to focus on the target object only, and improve the object reconstruction quality with the increasing number of views.

\begin{figure}[t!]
    \centering
    \begin{tabular}{cc}
        \includegraphics[width=1.0\columnwidth,trim={0 0 0 0},clip]{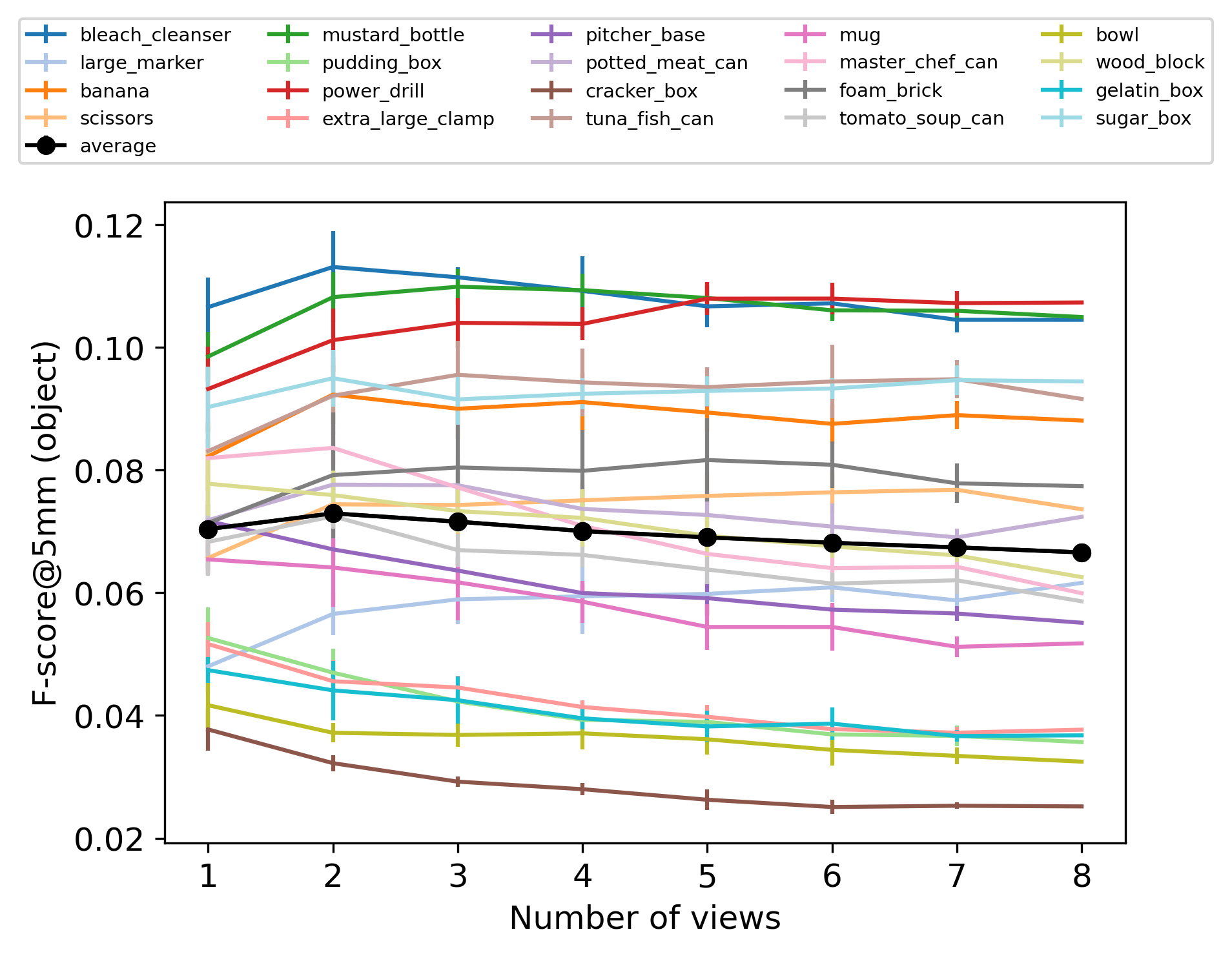}\\
    \end{tabular}
    \caption{Average F-score across 6 runs per object when varying the number of input views
    }
    \label{fig:multi_perobj}
    \vspace{-5pt}
\end{figure}

We further analyse the F-score per object with varying numbers of views in Figure~\ref{fig:multi_perobj}. We observe that the reconstruction quality of each object changes differently to the increasing number of views. Objects that have good reconstruction quality at a single-view benefits the most when the number of views increases, while objects that have lower reconstruction quality become worse when the number of views increases.

We also observe that objects that stand out from the background can benefit the most from the increasing number of views, for example mustard bottle, power drill and foam brick. This further supports the need for hand-object segmentation masks to allow our reconstruction model to fully take advantage of the multiple views when reconstructing the object.

\begin{figure*}[t!]
    \centering
    \begin{tabular}{c}
        \includegraphics[width=1.8\columnwidth]{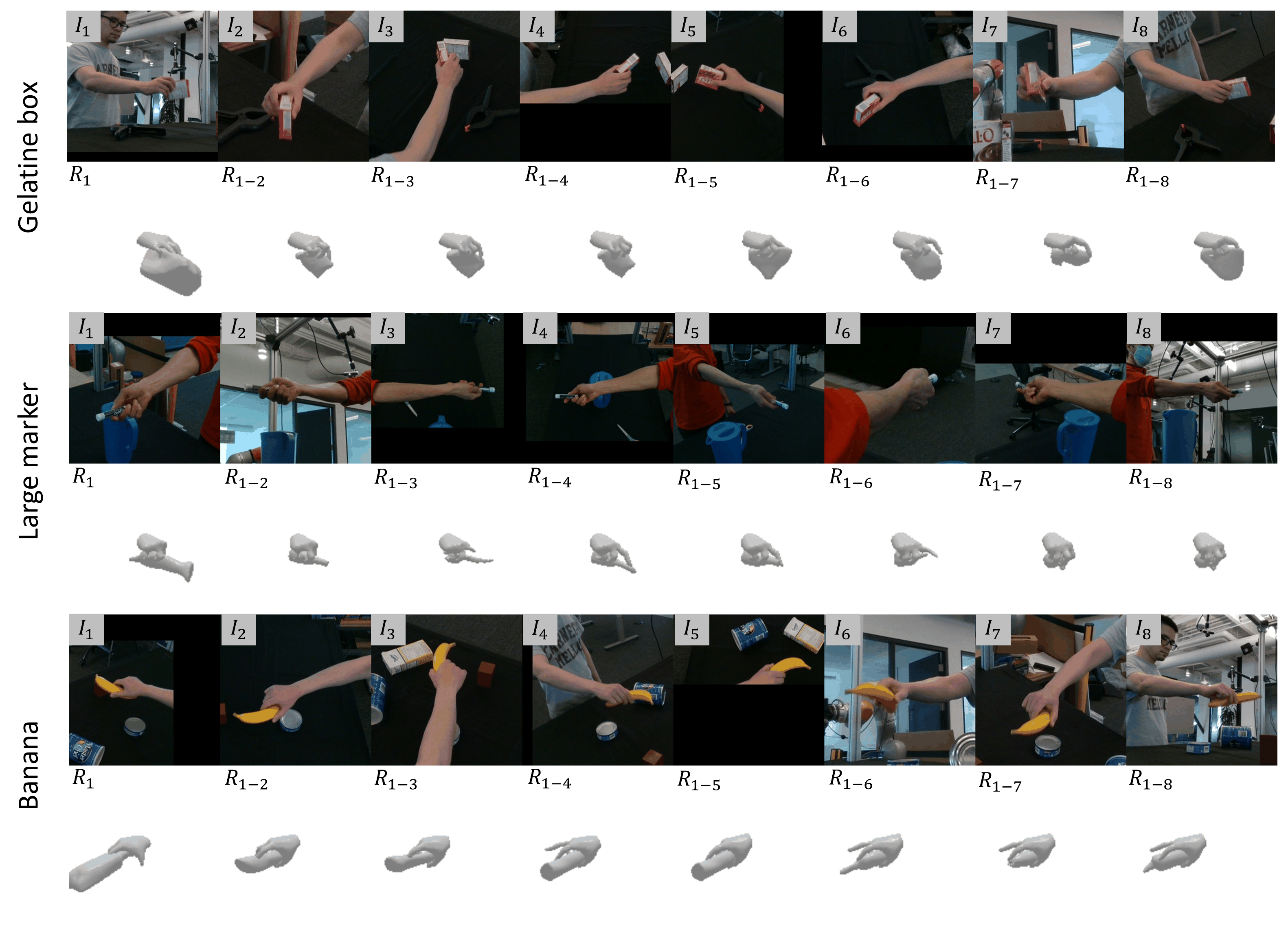}\\
    \end{tabular}
    \caption{Qualitative reconstruction results from multi-view RGB. The top row shows the 8 views used in the reconstruction. The bottom row shows (from left to right) the reconstruction obtained using 1 to 8 views.
    }
    \label{fig:vis_multi}
    \vspace{-5pt}
\end{figure*}

We show additional qualitative results in Figure~\ref{fig:vis_multi}. For the gelatine box in the first example, we observe that the first view used $I_1$ has a cluttered background with colours similar to that of the gelatine box. As the model receives the second view $I_2$ and third view $I_3$, the reconstruction improves as the object can be clearly separated from the dark background. However, further increasing the number of views causes the reconstruction to deteriorate, as more and more objects appear in the background. Similar issues can be observed in the fourth view $I_4$ and the sixth view $I_6$ of the large marker and the banana as the target object blends with the background or other objects in the scene. Self occlusion by the hand on the object can also cause the reconstruction to deteriorate especially for smaller objects.

\section{Conclusion}
In this paper, we tackled the challenging problem of sparse multi-view reconstruction of unseen (hand-held) objects. We showed that our model can take advantage of the additional information from multiple views for reconstruction. However, in cluttered scenes, increasing the number of views may negatively impact the reconstruction quality of the object. Our future work includes the introduction of a hand-object segmentation model as a pre-processing step for the model to better take advantage of the multiple views.

\section*{Acknowledgement}
This project made use of time on Tier 2 HPC facility JADE2, funded by EPSRC (EP/T022205/1).


{\small
\bibliographystyle{ieee_fullname}
\bibliography{egbib_short}
}

\end{document}